\documentclass[10pt,twocolumn,letterpaper]{article}

\usepackage[pagenumbers]{cvpr_2025} % To force page numbers, e.g. for an arXiv version

\definecolor{cvprblue}{rgb}{0.21,0.49,0.74}
\usepackage[pagebackref,breaklinks,colorlinks,allcolors=cvprblue]{hyperref}
\usepackage{color}
\usepackage{multirow}
\usepackage[accsupp]{axessibility}
\usepackage{makecell}
\usepackage{color}
\usepackage{colortbl}
\usepackage{bbding}
\definecolor{lightpink}{rgb}{1.0, 0.71, 0.76}
\definecolor{lightblue}{rgb}{0.68, 0.85, 0.9}
\usepackage{algorithm}
\usepackage{algorithmic}
\definecolor{mygray}{gray}{.93}

\def\paperID{5443} % *** Enter the Paper ID here
\def\confName{CVPR}
\def\confYear{2025}

\title{ReconDreamer: Crafting World Models for Driving Scene\\Reconstruction via Online Restoration}

\author{
~~~~~~Chaojun Ni\footnotemark[1]~\textsuperscript{\rm \ 1, 2}
~~~~~~Guosheng Zhao\footnotemark[1]~\textsuperscript{\rm \ 1, 4}
~~~~~~Xiaofeng Wang\footnotemark[1]~\textsuperscript{\rm \ 1, 4}
~~~~~~Zheng Zhu\footnotemark[1]~\textsuperscript{\rm \ 1}\textsuperscript{\Envelope}
~~~~~~Wenkang Qin\textsuperscript{\rm 1}\\
~~~~~~Guan Huang\textsuperscript{\rm 1} 
~~~~~~Chen Liu\textsuperscript{\rm 3} 
~~~~~~Yuyin Chen\textsuperscript{\rm 3}
~~~~~~Yida Wang\textsuperscript{\rm 3}
~~~~~~Xueyang Zhang\textsuperscript{\rm 3} 
~~~~~~Yifei Zhan\textsuperscript{\rm 3} \\
~~~~~~Kun Zhan\textsuperscript{\rm 3} 
~~~~~~Peng Jia\textsuperscript{\rm 3} 
~~~~~~Xianpeng Lang\textsuperscript{\rm 3} 
~~~~Xingang Wang\textsuperscript{\rm 4}
~~~~Wenjun Mei\textsuperscript{\rm 2}\textsuperscript{\Envelope}
\\
\textsuperscript{\rm 1}GigaAI
~ ~ \textsuperscript{\rm 2}Peking University
~ ~ \textsuperscript{\rm 3}
Li Auto Inc.
~ ~ \textsuperscript{\rm 4}CASIA
\\
\small{Project Page: \url{https://recondreamer.github.io}}
}

\begin{document}
\twocolumn[{%
\vspace{-1em}
\maketitle
\vspace{-2em}
\begin{center}

\centering
\setlength{\abovecaptionskip}{0.5em}
\resizebox{0.97\linewidth}{!}{
\includegraphics{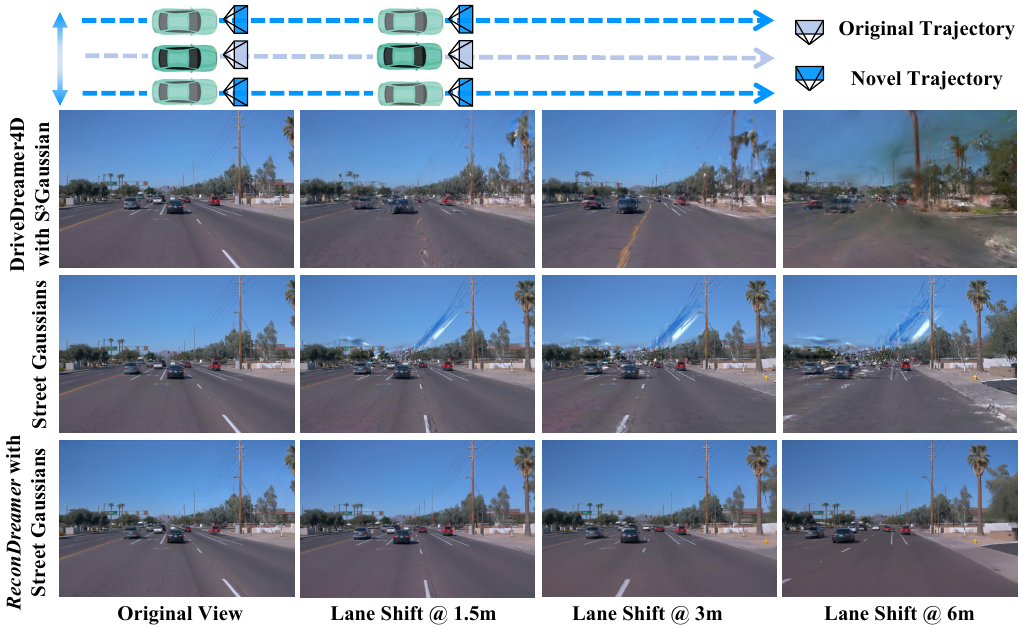
}}
\captionof{figure}{
Dynamic driving scene reconstruction methods, such as DriveDreamer4D \cite{drivedreamer4d} and Street Gaussians \cite{streetgaussian}, encounter significant challenges when rendering larger maneuvers (e.g., multi-lane shifts). In contrast, the proposed \textit{ReconDreamer} significantly improves rendering quality via incrementally integrating world model knowledge.
}
\label{fig:main}
% \vspace{-0.5em}
\end{center}}]

\renewcommand{\thefootnote}{\fnsymbol{footnote}}
\footnotetext[1]{These authors contributed equally to this work.
\textsuperscript{\Envelope}Corresponding authors: Zheng Zhu, zhengzhu@ieee.org, Wenjun Mei, mei@pku.edu.cn.}

%%%%%%%%% ABSTRACT
\begin{abstract}

Closed-loop simulation is crucial for end-to-end autonomous driving. Existing sensor simulation methods (e.g., NeRF and 3DGS) reconstruct driving scenes based on conditions that closely mirror training data distributions. However, these methods struggle with rendering novel trajectories, such as lane changes. 
Recent works have demonstrated that integrating world model knowledge alleviates these issues. Despite their efficiency, these approaches still encounter difficulties in the accurate representation of more complex maneuvers, with multi-lane shifts being a notable example.
Therefore, we introduce \textit{ReconDreamer}, which enhances driving scene reconstruction through incremental integration of world model knowledge. Specifically, \textit{DriveRestorer} is proposed to mitigate artifacts via online restoration. This is complemented by a progressive data update strategy designed to ensure high-quality rendering for more complex maneuvers. To the best of our knowledge, \textit{ReconDreamer} is the first method to effectively render in large maneuvers. Experimental results demonstrate that \textit{ReconDreamer} outperforms Street Gaussians in the NTA-IoU, NTL-IoU, and FID, with relative improvements by 24.87\%, 6.72\%, and 29.97\%. Furthermore, \textit{ReconDreamer} surpasses DriveDreamer4D with PVG during large maneuver rendering, as verified by a relative improvement of 195.87\% in the NTA-IoU metric and a comprehensive user study.

\end{abstract}

%\vspace{-0.7cm}
\section{Introduction}
Open-loop simulation techniques have made significant advancements in the field of autonomous driving \cite{stp3, uniad, vad}. However, current open-loop evaluation methods fall short in providing an accurate assessment of end-to-end planning algorithms, underscoring the need for more robust evaluation frameworks \cite{li2024ego, admlp,xld}. A promising approach to address this issue involves using closed-loop evaluations conducted in real-world scenarios, which require retrieving sensor data from novel trajectory views. This demands the driving scene representation capable of reconstructing the intricate and dynamic nature of driving environments.

Closed-loop simulation predominantly hinges on scene reconstruction approaches like Neural Radiance Fields (NeRF) \cite{nerf,emernerf,unisim,streetsurf} and 3D Gaussian Splatting (3DGS) \cite{3dgs,streetgaussian,s3gaussian,omnire}. Despite their contributions, these techniques are fundamentally restricted by the density and diversity of training data, often limiting their rendering capabilities to scenarios that closely mimic the original training data. Consequently, they underperform in complex, high-variation driving maneuvers. Current developments in autonomous driving world models \cite{drivedreamer,drivedreamer2,worlddreamer,gaia,drivewm,vista} have introduced the capability to generate diverse videos aligned with specific driving commands, renewing the potential for more robust closed-loop simulation. The recent DriveDreamer4D \cite{drivedreamer4d} has further evidenced that leveraging pretrained world models as data machines can substantially improve the quality of dynamic driving scene reconstruction. However, while this training-free integration of world model knowledge is efficient, its current design still encounters challenges in executing larger maneuvers (e.g., multi-lane shifts).

In this paper, we introduce \textit{ReconDreamer}, which enhances driving scene reconstruction via incrementally integrating knowledge from autonomous driving world models. Unlike \cite{drivedreamer4d} which leverages pretrained world models to directly expand novel trajectory views, \textit{ReconDreamer} trains the world model to progressively mitigate ghosting artifacts in complex maneuver renderings. Specifically, we generate a video restoration dataset by sampling rendering outputs at various training stages. Based on the dataset, we propose the \textit{DriveRestorer}, which is fine-tuned upon the world model to mitigate ghosting artifacts via online restoration. During the training, the masking strategy is introduced to emphasize restoration of challenging areas (e.g., sky and distant regions). Furthermore, we propose the Progressive Data Update Strategy (PDUS) to gradually restore artifacts, which ensures high-quality rendering for larger maneuvers. The proposed PDUS, by incrementally integrating world model knowledge, reduces the complexity of video restoration, making \textit{ReconDreamer} the first approach capable of handling large viewpoint shifts in rendering (e.g., across multiple lanes, spanning up to 6 meters).
As illustrated in Fig.~\ref{fig:main}, experimental results confirm that \textit{ReconDreamer} substantially improves Street Gaussians \cite{streetgaussian} during novel trajectory rendering, achieving a relative improvement in the average NTA-IoU, NTL-IoU, and FID by 24.87\%, 6.72\%, and 29.97\%. Additionally, \textit{ReconDreamer} strengthens spatiotemporal coherence in larger maneuvers, outperforming DriveDreamer4D \cite{drivedreamer4d} with a win rate 96.88\% in the user study, and a relative improvement of 195.87\% in the NTA-IoU metric.

The primary contributions of this work are as follows: (1) We present \textit{ReconDreamer}, which enhances dynamic driving scene reconstruction via incremental integration of world model knowledge. Notably, to our knowledge, \textit{ReconDreamer} is the first method to effectively render in large maneuvers (e.g., spanning up to 6 meters).
(2) The \textit{DriveRestorer} is proposed to mitigate ghosting artifacts via online restoration. Besides, we introduce the progressive data update strategy to maintain high-quality rendering for larger maneuvers.
(3) We perform comprehensive experiments to validate that \textit{ReconDreamer} can enhance rendering quality during large maneuvers, as well as the spatiotemporal coherence of driving scene elements. 

\begin{figure*}[!t]
\centering
\setlength{\abovecaptionskip}{0.5em}
\includegraphics[width=\textwidth]{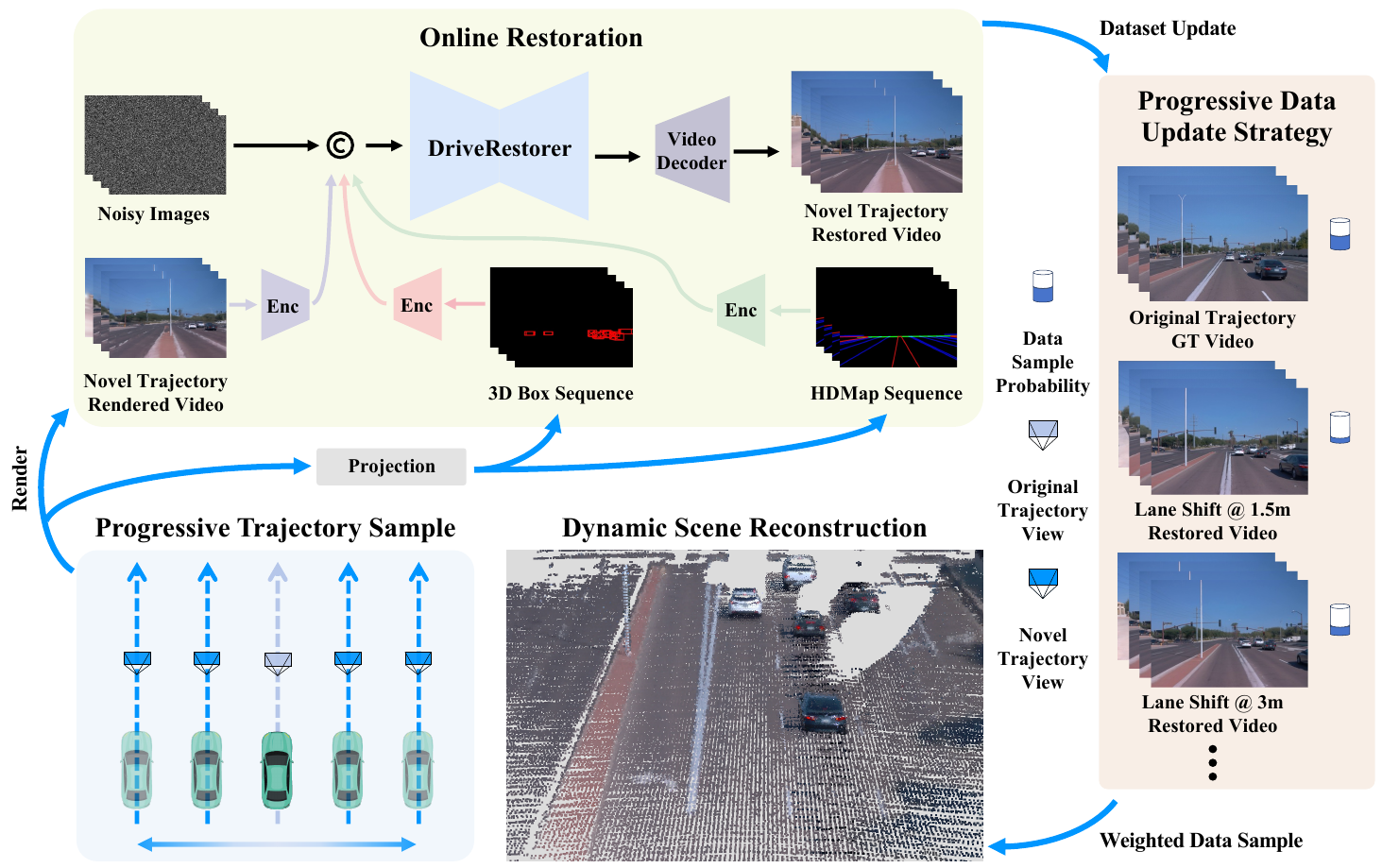}
\caption{The overall framework of \textit{ReconDreamer}. During the training of the dynamic driving scene reconstruction, we begin by rendering novel trajectory views. These rendered videos are subsequently processed by the \textit{DriveRestorer} to restore their quality. Then these restored videos, together with the original video, are employed to optimize the reconstruction model. This iterative process continues until the reconstruction model converges (The training of \textit{DriveRestorer} is omitted in this figure, and more details are in Sec.~\ref{sec:driveres}).}
\label{fig_ReconDreamer}
\vspace{-1.5em}
\end{figure*}

\section{Related Work}
\subsection{Driving Scene Reconstruction Methods}
NeRF and 3DGS have become prominent techniques in scene reconstruction. NeRF models \cite{nerf,mipnerf,zipnerf,ngp} use multi-layer perceptron (MLP) networks to represent continuous volumetric scenes, achieving exceptional rendering quality. Recently, 3DGS \cite{3dgs,mipgs} introduced a novel approach by defining anisotropic Gaussians in 3D space and employing adaptive density control, which allows for high-quality renderings even from sparse point cloud data.
Various studies have adapted NeRF \cite{unisim,emernerf,neo360,lu2023urban,urbannerf,blocknerf,streetsurf,ucnerf} and 3DGS \cite{streetgaussian,drivinggaussian,pvg,sgd,s3gaussian,omnire,hogaussian} for driving scene reconstructions. To accommodate the dynamic nature of driving environments, some methods incorporate time as an additional parameter to capture temporal variations in dynamic scenes \cite{attal2023hyperreel,kplane,li2021neural,lin2022efficient,hypernerf,nerfplayer,s3gaussian}, while others treat the scene as a combination of moving objects overlaid on a static background \cite{unisim,panopticnerf,ost2021neural,Neurad,mars,snerf}. Despite these advances, NeRF and 3DGS-based methods still encounter challenges related to data density. Their effectiveness in rendering relies heavily on sensor trajectory aligning closely with the training distribution. To tackle these challenges, methods like SGD \cite{sgd}, GGS \cite{han2024ggs}, and MagicDrive3D \cite{magicdrive3d} utilize generative models to expand the diversity of training perspectives. However, these methods are primarily focused on sparse image data or static background components, making them insufficient to fully capture the complexities of dynamic driving environments.

\subsection{World Models}
World models predict possible future world states as a function of imagined action sequences proposed by the actor \cite{lecun2022jepa,zhu2024sora}. 
Based on world models, recent methods \cite{worlddreamer,yan2021videogpt,pixeldance,emuvideo,gupta2023photorealistic,svd,ho2022imagen,ma2024latte,ho2022video,videoldm,kondratyuk2023videopoet,yang2024cogvideox,hong2022cogvideo,xiang2024pandora,egovid} have advanced the simulation of environments by generating videos that are guided by free-text actions. Leading this development is Sora \cite{videoworldsimulators2024}, which employs cutting-edge generative methods to create complex visual sequences that adhere to the physical laws governing the environment. This capability not only enhances the fidelity of generated video content but also holds significant potential for applications in real-world driving scenarios. In autonomous driving, world models \cite{drivedreamer,drivedreamer2,drivewm,gaia,vista,yang2024drivearena} utilize predictive techniques to interpret driving environments. These methods generate realistic driving scenarios while extracting driving policies from video data, renewing the potential for more robust closed-loop simulations. The recent DriveDreamer4D \cite{drivedreamer4d} has further evidenced that leveraging pretrained world models as data machines improves dynamic driving scene reconstruction. 
Nonetheless, it still encounters challenges in executing larger maneuvers (e.g., multi-lane shifts).

% \subsection{Generative Prior for Scene Reconstruction}
% Scene reconstruction from limited observations demands generative prior, particularly for unseen areas. Earlier studies distill the knowledge from text-to-image models \cite{sd,sdxl,saharia2022photorealistic,ramesh2022hierarchical} into a 3D representation model. Specifically, the Score Distillation Sampling (SDS) \cite{dreamfusion,magic3d,reconfusion} is adopted to synthesize a 3D object from the text prompt. Furthermore, to enhance 3D consistency, several approaches extend the multi-view diffusion models \cite{sargent2023zeronvs,gao2024cat3d} and video diffusion models \cite{svd,voleti2024sv3d,chen2024v3d} to 3D scene generation. 
% To extend the diffusion prior to complex, dynamic, large-scale driving scenes for 3D reconstruction, methods such as SGD \cite{sgd}, GGS \cite{han2024ggs} and MagicDrive3D \cite{magicdrive3d} employ generative models to broaden the range of training viewpoints. 

\section{Method}
%In this section, the overall framework of \textit{ReconDreamer} is first described. Then we elaborate on training and inference details of the \textit{DriveRestorer}, along with the progressive data update strategy.

\begin{figure}[!t]
\centering
\setlength{\abovecaptionskip}{0.5em}
\includegraphics[width=0.45\textwidth]{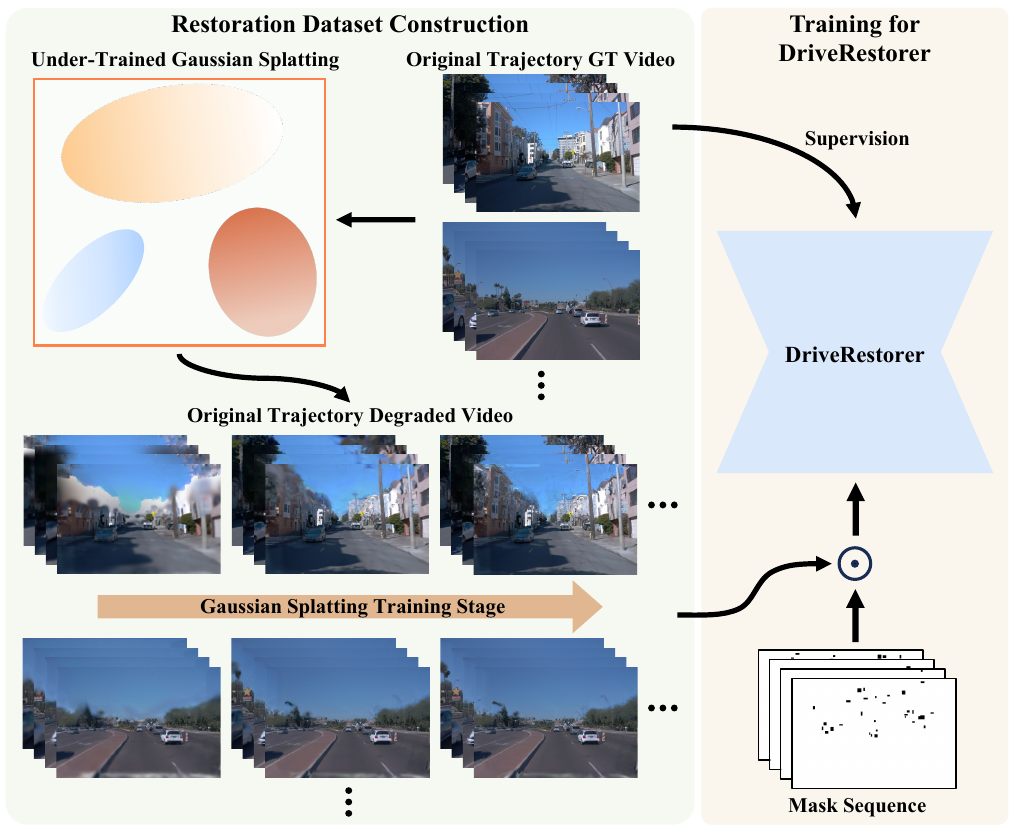}
\caption{The restoration dataset construction for training \textit{DriveRestorer}. Initially, we train an under-trained Gaussian Splatting model utilizing the ground truth (GT) videos from the original trajectory. During training of Gaussian Splatting model, degraded videos of the original trajectory are rendered at each stage. These degraded videos, paired with their corresponding GT videos, form the restoration dataset. A mask is then applied to the degraded videos to train DriveRestorer, supervised by the GT videos. }
\label{fig_DriveRestorer}
\vspace{-1.5em}
\end{figure}

\subsection{Overall Framework of ReconDreamer}
Traditional scene reconstruction methods \cite{3dgs,nerf,pvg,streetgaussian,s3gaussian,4dgs,deformablegs} face challenges due to the sparsity of training data. Recent approaches \cite{sgd,han2024ggs,drivedreamer4d} alleviate this issue by leveraging generative priors to increase the data density. However, a gap remains between the generated data and the real data. In contrast, the proposed \textit{ReconDreamer} expands the training data through an online restoration process. Notably, \textit{ReconDreamer} progressively restores rendered data, effectively reducing the gap between generated and original data. 
The overall framework of \textit{ReconDreamer} is illustrated in Fig.~\ref{fig_ReconDreamer}. Specifically, we first train the scene reconstruction method $\mathcal{G}$ using the original data $V_{\text{ori}}$. The trained methods then render novel trajectory videos $\hat{V}_{\text{novel}}$:
\begin{equation}
    \hat{V}_{\text{novel}} = \mathcal{G}(\mathcal{T}_{\text{novel}}),
\end{equation}
where $\mathcal{T}_{\text{novel}}$ is the novel trajectory. Notably, $\hat{V}_{\text{novel}}$ exhibits ghost artifacts due to data sparsity. Therefore, the \textit{DriveRestorer} is introduced to restore the artifacts. The restoration resembles the diffusion denoising process \cite{drivedreamer2}, where structure conditions (3D boxes, and HDMaps) are employed to ensure spatiotemporal coherence of traffic elements. Note that the conditions are processed via projection transformation to align with $\mathcal{T}_{\text{novel}}$ (see Sec.~\ref{sec:driveres} for more details). Consequently, the restored renderings $V_{\text{novel}}$ have a smaller gap with the original video $V_{\text{ori}}$, making them more suitable as training samples for scene reconstruction. 
Additionally, to further enhance the training of $\mathcal{G}$ and enable it to render large maneuvers (e.g., multi-lane shifts), we propose the PDUS, which progressively updates the training dataset for scene reconstruction. Specifically, the novel trajectory is gradually expanded to generate large maneuver videos. These videos are then restored by \textit{DriveRestorer} and used to update the training dataset (see Sec.~\ref{sec:pdus} for more details). The updated dataset is then employed to optimize the reconstruction model. This iterative process continues until the reconstruction model converges.

% gradually sample restorations at larger maneuver $\mathcal{T}_\text{large}^k$:
% \begin{equation}
%     V_{\text{novel}}=\mathcal{R}(\mathcal{G}(\mathcal{T}_\text{large}^k), \mathcal{P}(c, \mathcal{T}_\text{large}^k)),
% \end{equation}
% where $\mathcal{T}_\text{large}^k$ is gradually expanded to reduces the restoration difficulty.
% Finally, the restored renderings $V_\text{novel}$ are fused with original data under different sampling probability, enriching the training data and progressively enhancing scene reconstruction under large maneuver conditions.

% \begin{figure}[!t] % 使用普通的 figure 环境
% \centering
% \includegraphics[width=\columnwidth]{figures/res_v3.pdf} % \columnwidth 代表一栏的宽度
% \caption{} % 添加你的图注
% \label{fig_ReconDreamer} % 图片标签
% \end{figure}

% \begin{figure}[!t] % 使用普通的 figure 环境
% \centering
% \includegraphics[width=\columnwidth]{figures/dataset.pdf} % \columnwidth 代表一栏的宽度
% \caption{} % 添加你的图注
% \label{fig_ReconDreamer} % 图片标签
% \end{figure}

\begin{figure}[!t]
\centering
\setlength{\abovecaptionskip}{0.5em}
\includegraphics[width=0.49\textwidth]{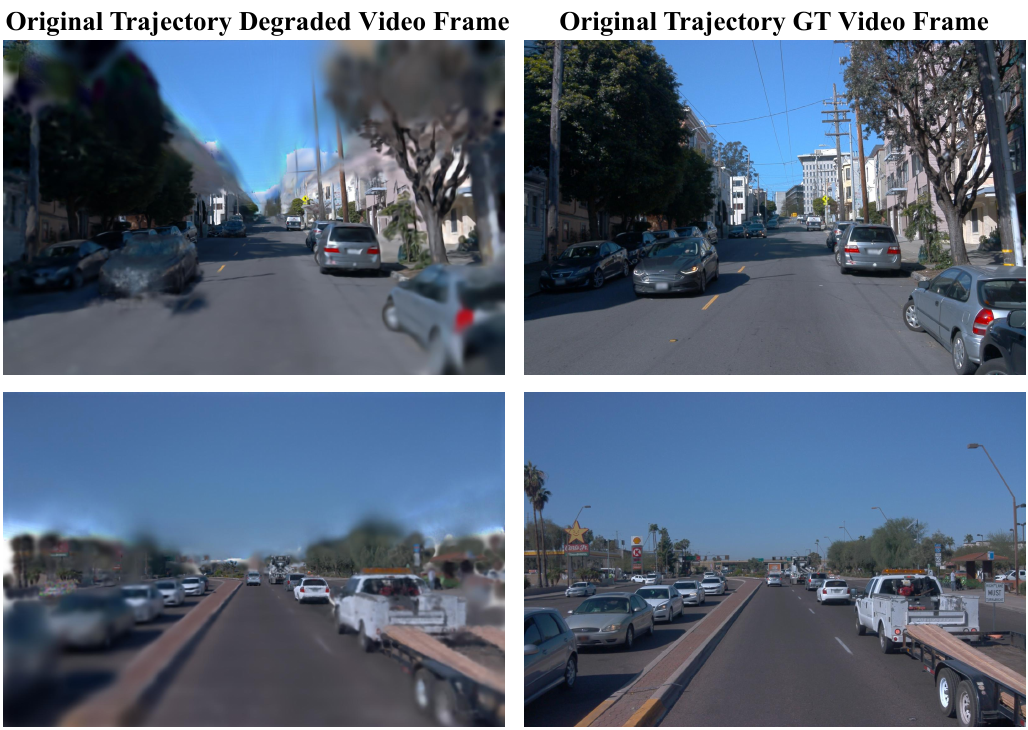}
\caption{Restoration data pairs for \textit{DriveRestorer} training.}
\label{fig:dataset} % 图片标签
\vspace{-1.5em}
\end{figure}

\subsection{Training and Inference of DriveRestorer}
\label{sec:driveres}
Traditional scene reconstruction methods often suffer from artifacts when rendering novel trajectory views. To address this issue, we introduce \textit{DriveRestorer} to restore these degraded renderings. In the next, we elaborate on training and inference details of the \textit{DriveRestorer}.

\noindent
\textbf{Training.}
The major challenge of training \textit{DriveRestorer} lies in the absence of rendering restoration datasets. Therefore, we propose a novel method for constructing restoration pairs. As illustrated in Fig.~\ref{fig_DriveRestorer}, we leverage under-trained reconstruction models \cite{pvg, deformablegs, s3gaussian, streetgaussian} to render videos $\hat{V}_{\text{ori}}=\mathcal{G}(\mathcal{T}_\text{ori})$ along the original trajectory, naturally producing ghosting artifacts due to model underfitting. These degraded frames are then paired with their corresponding ground truth video $V_{\text{ori}}$, forming a rendering restoration dataset. To further enhance dataset diversity, we sample rendered videos from different training stages. Consequently, the constructed rendering restoration pairs are represented as $\{\hat{V}_{\text{ori}}^k, V_{\text{ori}}\}$, where $\hat{V}_{\text{ori}}^k$ denotes the degraded video frame sampled at training stage $k$. Fig.~\ref{fig:dataset} provides a visualization of these data pairs.
Based on the constructed dataset, we train \textit{DriveRestorer} to restore artifacts in the rendered videos. The \textit{DriveRestorer} is fine-tuned upon the world model \cite{drivedreamer2}. Specifically, we introduce degraded video frames $\hat{V}_{\text{ori}}$ as a control condition to provide appearance priors. To further emphasize the restoration of challenging regions, we apply masks to the degraded video frames $\hat{V}_{\text{ori}}$ during training. Since video quality degrades in distant areas (further from the camera center) and at the sky-scene boundary, our masks $M$ focus primarily on these problematic regions (see the supplement for more details).
During the training of \textit{DriveRestorer}, we first feed the masked video frame $\hat{V}_{\text{mask}} = \hat{V}_{\text{ori}} \odot M$ into the encoder $\mathcal{E}$ to obtain the low-dimensional latent feature $z = \mathcal{E}(\hat{V}_{\text{mask}})$. The fine-tuning process of the world model is optimized using the diffusion loss:
\begin{equation}
    \mathcal{L}_{\mathcal{R}} = \mathbb{E}_{\boldsymbol{z}, \epsilon \sim \mathcal{N}(0,1), t}\left[\left\|\epsilon_{t}-\epsilon_{\theta}\left(\boldsymbol{z}_{t}, t, \boldsymbol{c}\right)\right\|_{2}^{2}\right],
\end{equation}
where $\epsilon_t$ denotes the random noise at time step $t$, $\epsilon_\theta$ is the parameterized denoising network, $\boldsymbol{z}_{t}$ refers to the noisy latent at time step $t$, and $c$ represents the control conditions, including the degraded video $\hat{V}_{\text{mask}}$, 3D boxes, and HDMaps. The integration of video, 3D boxes, and HDMaps is similar to that of \cite{drivedreamer2}, with further details provided in the supplement.

\begin{figure}[!t]
\centering
\setlength{\abovecaptionskip}{0.5em}
\includegraphics[width=0.45\textwidth]{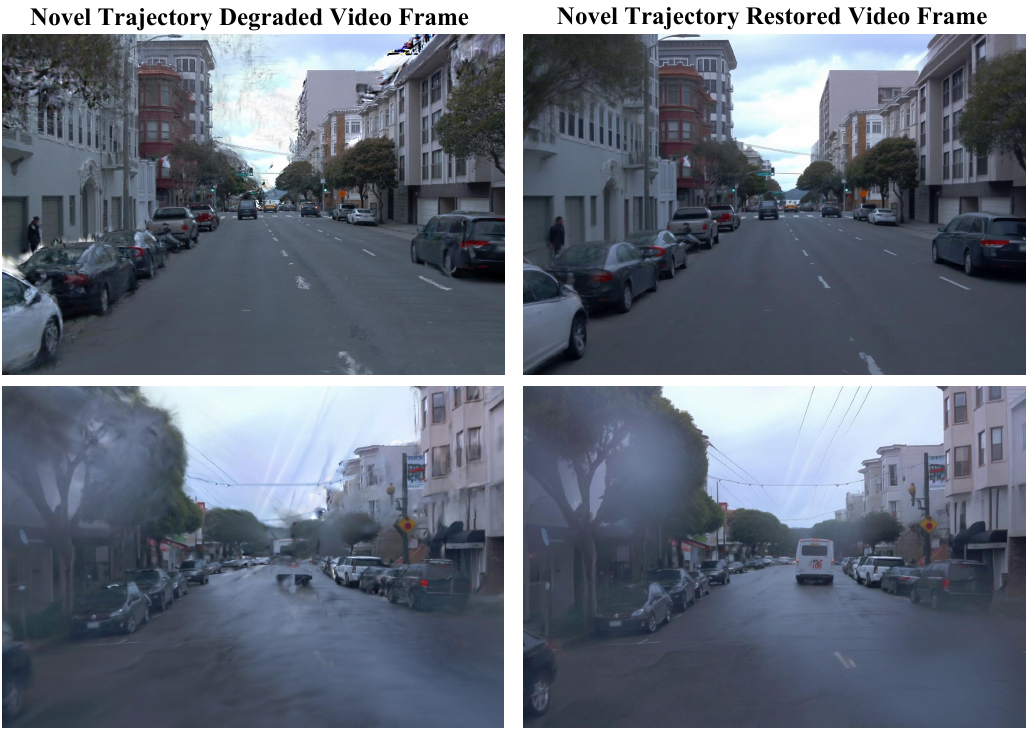}
\caption{Examples of degraded video frame rendered under new trajectories and their restored frame by \textit{DriveRestorer}.}
\label{fig:res_v6} % 图片标签
\vspace{-1.5em}
\end{figure}

\noindent
\textbf{Inference.} After training \textit{DriveRestorer}, we freeze its parameters to restore novel trajectory renderings:
\begin{equation}
    V_\text{novel} = \mathcal{R}(\hat{V}_{\text{novel}}, \mathcal{P}(s,\mathcal{T}_{\text{novel}}^k)),
\end{equation}
where $\mathcal{R}$ is the \textit{DriveRestorer}, $s$ is the structure conditions (3D boxes and HDMaps),
$\mathcal{P}(\cdot)$ denotes the projection transformation that aligns structure conditions with $\mathcal{T}_\text{novel}^k$. Note that $\mathcal{T}_\text{novel}^k$ is progressively expanded at different training stage $k$, which enhances driving scene reconstruction under large maneuver conditions (refer to Sec.~\ref{sec:pdus} for more details).
As shown in Fig.~\ref{fig:res_v6}, the trained \textit{DriveRestorer} can mitigate ghost artifacts in novel trajectory renderings.

\vspace{-0.5em}
\begin{algorithm}
    \caption{Progressive Data Update Strategy}
    \label{alg_pdus}
    \renewcommand{\algorithmicrequire}{\textbf{Input:}}
    \renewcommand{\algorithmicensure}{\textbf{Output:}}
    
    \begin{algorithmic}
        \REQUIRE Reconstruction model $\mathcal{G}$, \textit{DriveRestorer} $\mathcal{R}$, Training step $T$, Update step $S$, Stride $\Delta y$, Original trajectory $\mathcal{T}_\text{ori}$, Novel trajectory video dataset $D_\text{novel}$
        \ENSURE Updated novel trajectory video dataset $D_\text{novel}$
        \STATE $k\gets 0$ 
        \FOR{$t$ in Range(0, $T$, $S$)}
        \STATE $k \gets k+1$ 
        \STATE $w \gets k /\sum_{j=1}^k j$
        \STATE $\mathcal{T}_\text{novel}^k$.y = $\mathcal{T}_\text{ori}$.y + $k\Delta y$ 
        % \vspace{0.3em}
        \STATE $\hat{V}_\text{novel} \gets \mathcal{G}(\mathcal{T}_\text{novel}^k)$
        \STATE $V_\text{novel} \gets \mathcal{R}(\hat{V}_\text{novel})$
        \STATE $D_\text{novel} \gets (1-w)\cdot D_\text{novel} \cup w\cdot V_\text{novel}$
        \ENDFOR
    \end{algorithmic}
\end{algorithm}
\vspace{-0.5em}

\subsection{Progressive Data Update Strategy}
\label{sec:pdus}

Based on \textit{DriveRestorer}'s capability to restore novel trajectory videos, we propose the Progressive Data Update Strategy (PDUS) to enhance driving scene reconstruction under large maneuver conditions. The PDUS first constructs a mixed dataset $ D = 0.5D_\text{ori} \cup 0.5D_\text{novel}$, where $D_\text{ori}$ is the original trajectory video dataset and $D_\text{novel}$ refers to the restored novel trajectory video dataset, which can be updated throughout the training process. The update strategy, detailed in Algo.~\ref{alg_pdus}, uses an update distance of $ y = k\Delta y$ meters at $k$-th update step to progressively update the novel trajectory $\mathcal{T}_{\text{novel}}$. Then the reconstruction model $\mathcal{G}$ renders novel trajectory video $\hat{V}_\text{novel}$, which are then processed by \textit{DriveRestorer} to obtain the restored novel trajectory video $V_\text{novel}$. To ensure that newly generated data provides additional priors for the reconstruction model, the updated dataset $D_\text{novel}$ can be obtained as follows:
\begin{equation}
    D_\text{novel}=(1-w) \cdot D_\text{novel} \cup w \cdot V_\text{novel},
\end{equation}
where $w=\frac{k}{\sum_{{j}=1}^{k}j}$ is the sampling probability for $V_\text{novel}$.

For Gaussian Splatting-based driving scene reconstruction methods, we use original trajectory dataset $D_\text{ori}$ to train the Gaussian parameters $\phi$:
\begin{equation}
    \mathcal{L}_\text{ori}(\phi) =\lambda_1\mathcal{L}_\text{ori}^\text{RGB}+\lambda_2\mathcal{L}_\text{ori}^\text{Depth} +\lambda_3\mathcal{L}_\text{ori}^\text{SSIM},
\end{equation}
where $\mathcal{L}_\text{ori}^\text{RGB}, \mathcal{L}_\text{ori}^\text{Depth}, \mathcal{L}_\text{ori}^\text{SSIM}$ are reconstruction losses typically used in the Gaussian Splatting optimization \cite{streetgaussian}, and $\lambda_1,\lambda_2,\lambda_3$ are loss weights. Additionally, for the generated novel trajectory data $D_\text{novel}$, we adopt the depth-free training strategy from \cite{drivedreamer4d} to train \textit{ReconDreamer}:
\begin{equation}
    \mathcal{L}_\text{novel}(\phi)=\lambda_1\mathcal{L}_\text{novel}^\text{RGB} +\lambda_3\mathcal{L}_\text{novel}^\text{SSIM}.
\end{equation}
% The overall loss function for mixed training is defined as follows:
The overall loss function for mixed training is:
\begin{equation}
    \mathcal{L}(\phi)=\mathcal{L}_\text{ori}+\mathcal{L}_\text{novel}.
\end{equation}
\vspace{-2em}

\begin{table*}[t]
\centering
\setlength{\abovecaptionskip}{0.5em}
\resizebox{1\linewidth}{!}{
\begin{tabular}{@{}m{3.7cm}ccccccccccccc@{}}
\toprule
\multirow{2}{*}{Method} & \multicolumn{3}{c}{Lane Shift @ 3m} & \multicolumn{3}{c}{Lane Shift @ 6m} & \multicolumn{3}{c}{ Lane Change} &\multicolumn{3}{c}{Average} \\ \cmidrule(lr){2-4} \cmidrule(lr){5-7} \cmidrule(lr){8-10} \cmidrule(lr){11-13}
 & NTA-IoU $\uparrow$ & NTL-IoU $\uparrow$ & FID$\downarrow$ & NTA-IoU $\uparrow$ & NTL-IoU $\uparrow$ & FID$\downarrow$  & NTA-IoU $\uparrow$ & NTL-IoU $\uparrow$ & FID$\downarrow$  & NTA-IoU $\uparrow$ & NTL-IoU $\uparrow$  & FID$\downarrow$  \\ \midrule
% \makecell[l]{\textit{DriveDreamer4D} \\ with PVG\cite{drivedreamer4d}} & 0.340 & 51.32 & 129.05 & 0.121 & 49.28 & 210.37 & 0.428 & 53.78 & 76.24 & 0.296 & 51.46 & 138.55 \\
% \midrule
PVG \cite{pvg}& 0.242 & 49.99 & 134.74 & 0.118 & 48.32 & 212.19 & 0.256 & 50.70 & 105.29 & 0.205 & 49.67 & 150.74\\
$\text{S}^3$Gaussian \cite{s3gaussian} & 0.059 & 48.24&234.54 & 0.014 & 47.69&311.59 & 0.175 & 49.05& 124.90 & 0.083&  48.33 & 223.68\\
% \makecell[l]{\textit{DriveDreamer4D} \\ with $\text{S}^3$Gaussian\cite{drivedreamer4d}}  & 0.263 & 52.04 & 173.49 & 0.031 & 48.04 & 303.95 & 0.491 & 53.36 & 76.54 & 0.262 & 51.15 & 184.66\\
% \midrule
Deformable-GS \cite{deformablegs}  & 0.142 & 50.63 & 161.20 & 0.069 & 48.86 &260.83& 0.240 & 51.62 &92.34& 0.150 & 50.37 & 171.46\\
% \makecell[l]{\textit{DriveDreamer4D} \\ with Deformable-GS\cite{drivedreamer4d}} & 0.181 & 50.72 & 169.79 & 0.078 & 48.90 & 259.19 & 0.322 & 52.90 & 92.34 & 0.193 & 50.84 & 173.77\\
% \midrule
Street Gaussians \cite{streetgaussian} & 0.498 & 53.19  & 130.75 & 0.374 & 49.27 & 213.04 & 0.496 & 56.03 &86.46& 0.456 & 52.83 & 143.42\\
\midrule
\textit{ReconDreamer} with & \multirow{2}{*}{\textbf{0.539}} & \multirow{2}{*}{\textbf{54.58}} & \multirow{2}{*}{\textbf{93.56}}  & \multirow{2}{*}{\textbf{0.467}} & \multirow{2}{*}{\textbf{52.58}} & \multirow{2}{*}{\textbf{149.19}} & \multirow{2}{*}{\textbf{0.554}} & \multirow{2}{*}{\textbf{56.63}} &\multirow{2}{*}{\textbf{73.91}}  & \multirow{2}{*}{\textbf{0.517}} & \multirow{2}{*}{\textbf{54.60}} & \multirow{2}{*}{\textbf{105.55}}\\
Street Gaussians & & & & & & & & & &  &  &   \\
\bottomrule
\end{tabular}}
\caption{Comparison of NTA-IoU, NTL-IoU, and FID scores for different novel trajectory views with various methods \cite{s3gaussian,deformablegs,pvg,streetgaussian}.}
\label{tab:method_comparison}
\vspace{-0.5em}
\end{table*}

\begin{table*}[t]
\centering
\setlength{\abovecaptionskip}{0.5em}
\resizebox{1\linewidth}{!}{
\begin{tabular}{@{}m{3.4cm}ccc ccc ccc ccc c@{}}
\toprule
\multirow{2}{*}{Method} & \multicolumn{3}{c}{Lane Shift @ 3m} & \multicolumn{3}{c}{Lane Shift @ 6m} & \multicolumn{3}{c}{ Lane Change} &\multicolumn{3}{c}{Average} \\ \cmidrule(lr){2-4} \cmidrule(lr){5-7} \cmidrule(lr){8-10} \cmidrule(lr){11-13}
 & NTA-IoU $\uparrow$ & NTL-IoU $\uparrow$ & FID$\downarrow$ & NTA-IoU $\uparrow$ & NTL-IoU $\uparrow$ & FID$\downarrow$  & NTA-IoU $\uparrow$ & NTL-IoU $\uparrow$ & FID$\downarrow$  & NTA-IoU $\uparrow$ & NTL-IoU $\uparrow$  & FID$\downarrow$  \\ \midrule 
 PVG \cite{pvg} & 0.242 & 49.99 & 134.74 & 0.118 & 48.32 & 212.19 & 0.256 & 50.70 & 105.29 & 0.205 & 49.67 & 150.74\\
\rowcolor{mygray}
\makecell[l]{ DriveDreamer4D\\ with PVG \cite{drivedreamer4d}}  & \raisebox{-4pt}{0.340}  & \raisebox{-4pt}{51.32} & \raisebox{-4pt}{129.05} & \raisebox{-4pt}{0.121} & \raisebox{-4pt}{49.28} & \raisebox{-4pt}{210.37} & \raisebox{-4pt}{0.438} & \raisebox{-4pt}{53.06} & \raisebox{-4pt}{\textbf{71.52}} & \raisebox{-4pt}{0.300} & \raisebox{-4pt}{51.22} & \raisebox{-4pt}{136.98} \\
\textit{ReconDreamer}   &  \multirow{2}{*}{\textbf{0.474}}  &\multirow{2}{*}{\textbf{ 52.06}} & \multirow{2}{*}{\textbf{124.12}} & \multirow{2}{*}{\textbf{0.358}} & \multirow{2}{*}{\textbf{50.14}} & \multirow{2}{*}{\textbf{177.58}} & \multirow{2}{*}{\textbf{0.464}} & \multirow{2}{*}{\textbf{53.21}} & \multirow{2}{*}{74.32} & \multirow{2}{*}{\textbf{0.432}} & \multirow{2}{*}{\textbf{51.80}}&\multirow{2}{*}{\textbf{125.34}} \\
with PVG & & & & & & & & & &  &  &   \\
\midrule 
$\text{S}^3$Gaussian\cite{s3gaussian} & 0.059 & 48.24&234.54 & 0.014 & 47.69&311.59 & 0.175 & 49.05& 124.90 & 0.083&  48.33 & 223.68\\

\rowcolor{mygray}
\makecell[l]{DriveDreamer4D\\ with  S\textsuperscript{3}Gaussian \cite{drivedreamer4d} } &  \raisebox{-4pt}{0.263} & \raisebox{-4pt}{\textbf{51.07}} & \raisebox{-4pt}{173.49} & \raisebox{-4pt}{0.031} & \raisebox{-4pt}{48.04} & \raisebox{-4pt}{303.95} & \raisebox{-4pt}{\textbf{0.495}} & \raisebox{-4pt}{\textbf{53.42}} & \raisebox{-4pt}{\textbf{66.93}} & \raisebox{-4pt}{0.263} & \raisebox{-4pt}{\textbf{50.84}} & \raisebox{-4pt}{181.46} \\

\textit{ReconDreamer}    &  \multirow{2}{*}{\textbf{0.267}}  & \multirow{2}{*}{50.26} & \multirow{2}{*}{\textbf{110.41}} & \multirow{2}{*}{\textbf{0.177}} & \multirow{2}{*}{\textbf{50.54}} & \multirow{2}{*}{\textbf{152.89}} & \multirow{2}{*}{0.413} & \multirow{2}{*}{51.62} & \multirow{2}{*}{123.61} & \multirow{2}{*}{\textbf{0.286}} & \multirow{2}{*}{50.80} & \multirow{2}{*}{\textbf{128.97}} \\
with $\text{S}^3$Gaussian & & & & & & & & & &  &  &   \\

\midrule 
  Deformable-GS \cite{deformablegs} & 0.142 & 50.63 & 161.20 & 0.069 & 48.86 &260.83& 0.240 & 51.62 &92.34& 0.150 & 50.37 & 171.46\\
\rowcolor{mygray}
\makecell[l]{DriveDreamer4D\\ with Deformable-GS \cite{drivedreamer4d} \vspace{-13pt}}  & {0.181}  & {50.72} & {169.79} & {0.078} & {48.90} & {259.19} & {0.335} & {52.93} & {77.32} & {0.198} & {50.85} & {168.77} \\
% \makecell{DriveDreamer4D\\
% with Deformable-GS \cite{drivedreamer4d}}  & \multirow{2}{*}{0.181}  & \multirow{2}{*}{50.72} & \multirow{2}{*}{169.79} & \multirow{2}{*}{0.078} & \multirow{2}{*}{48.90} & \multirow{2}{*}{259.19} & \multirow{2}{*}{0.335} & \multirow{2}{*}{52.93} & \multirow{2}{*}{77.32} & \multirow{2}{*}{0.198} & \multirow{2}{*}{50.85} & \multirow{2}{*}{168.77} \\
% \rowcolor{mygray}
% with Deformable-GS \cite{drivedreamer4d} & & & & & & & & & &  &  &   \\
\textit{ReconDreamer}   &  \multirow{2}{*}{\textbf{0.416}}  & \multirow{2}{*}{\textbf{52.26}} & \multirow{2}{*}{\textbf{106.99}} & \multirow{2}{*}{\textbf{0.294}} & \multirow{2}{*}{\textbf{50.54}} & \multirow{2}{*}{\textbf{143.88}} & \multirow{2}{*}{\textbf{0.443}} & \multirow{2}{*}{\textbf{53.78}} & \multirow{2}{*}{\textbf{76.24}} & \multirow{2}{*}{\textbf{0.384}} & \multirow{2}{*}{\textbf{52.19}} & \multirow{2}{*}{\textbf{109.04}}\\
with Deformable-GS & & & & & & & & & &  &  &   \\

\bottomrule
\end{tabular}}
\caption{Comparison of NTA-IoU, NTL-IoU, and FID scores for different novel trajectory views with DriveDreamer4D \cite{drivedreamer4d}.}
\label{tab:dd4d_comparison}
\vspace{-1em}
\end{table*}

%%%%%%%%%%%%%%%%%%%%%%%%%%%%%%%%%%%%%%%%%%%%%%%%%%%%%%%%%%%%%%%%%%%%%%%%%%%%%%%
\begin{table}[t]
\centering
\setlength{\abovecaptionskip}{0.5em}
\resizebox{1\linewidth}{!}{
\begin{tabular}{l|cccc}
\toprule
\multirow{2}{*}{} & \multicolumn{2}{c}{\textit{ReconDreamer} Win Rate} \\& Street Gaussians\cite{streetgaussian} & DriveDreamer4D with PVG\cite{drivedreamer4d}
\\ \midrule
Lane Shift @ 3m & 97.92\% & 96.88\% \\
Lane Shift @ 6m & 100.00\% & 100.00\% \\
Lane Change &  93.75\% & 93.75\% \\
\midrule
\textbf{Average} & \textbf{97.22\%} & \textbf{96.88\%} \\
\bottomrule
\end{tabular}}
\caption{Comparing the win rates of \textit{ReconDreamer} in rendering large maneuvers.}
\label{tab:user}
\vspace{-2em}
\end{table}

\begin{table*}[t]
\centering
\setlength{\abovecaptionskip}{0.5em}
\resizebox{1\linewidth}{!}{
\begin{tabular}{@{}m{4cm}ccccccccccccc@{}}
\toprule
\multirow{2}{*}{Preatrained Model} & \multicolumn{3}{c}{Lane Shift @ 3m} & \multicolumn{3}{c}{Lane Shift @ 6m} & \multicolumn{3}{c}{Lane Change} & \multicolumn{3}{c}{Average} \\ \cmidrule(lr){2-4} \cmidrule(lr){5-7} \cmidrule(lr){8-10} \cmidrule(lr){11-13}
 & NTA-IoU $\uparrow$ & NTL-IoU $\uparrow$ & FID$\downarrow$ & NTA-IoU $\uparrow$ & NTL-IoU $\uparrow$ & FID$\downarrow$ & NTA-IoU $\uparrow$ & NTL-IoU $\uparrow$ & FID$\downarrow$ & NTA-IoU $\uparrow$ & NTL-IoU $\uparrow$ & FID$\downarrow$ \\ \midrule
Stable Diffusion \cite{sd} & 0.504 & 54.51 & 112.25 & 0.389 & 52.13 & 178.16 & 0.509 & 56.42 & 80.90 & 0.467 & 54.35 & 123.77 \\ % 使用 - 表示缺失值
Stable Video Diffusion\cite{svd} &0.500&53.33&120.46&0.435&49.67&199.40&0.527&56.05&89.89&0.487&53.02&136.58\\

DriveDreamer-2\cite{drivedreamer2} & 0.525 & 54.42 & 100.32 & 0.456 & 52.17 & 152.23 & \textbf{0.551} & \textbf{56.67} & 92.30 & 0.511 & 54.42 & 114.95 \\ 
DriveDreamer-2+Mask & \textbf{0.539} & \textbf{54.52} & \textbf{93.56} & \textbf{0.467} & \textbf{52.56} & \textbf{149.19} & 0.544 & \textbf{56.67} & \textbf{73.91} & \textbf{0.517} & \textbf{54.58} & \textbf{105.55} \\
\bottomrule
\end{tabular}}
\caption{Comparison of NTA-IoU, NTL-IoU, and FID scores for \textit{DriveRestorer} with different backbones.}
\label{tab:pretrained}
\vspace{-0.5em}
\end{table*}
% ##################
% ##################################################
%逐渐恢复表格
\begin{table*}[t]
\centering
\setlength{\abovecaptionskip}{0.5em}
\resizebox{1\linewidth}{!}{
\begin{tabular}{@{}m{1cm}ccccccccccccc@{}}
\toprule
\multirow{2}{*}{Stride} & \multicolumn{3}{c}{Lane Shift @ 3m} & \multicolumn{3}{c}{Lane Shift @ 6m} & \multicolumn{3}{c}{Lane Change} & \multicolumn{3}{c}{Average} \\ \cmidrule(lr){2-4} \cmidrule(lr){5-7} \cmidrule(lr){8-10} \cmidrule(lr){11-13}
 & NTA-IoU $\uparrow$ & NTL-IoU $\uparrow$ & FID$\downarrow$ & NTA-IoU $\uparrow$ & NTL-IoU $\uparrow$ & FID$\downarrow$ & NTA-IoU $\uparrow$ & NTL-IoU $\uparrow$ & FID$\downarrow$ & NTA-IoU $\uparrow$ & NTL-IoU $\uparrow$ & FID$\downarrow$ \\ \midrule
        0.5m & 0.533 & 54.10 & 120.23 & 0.452 & 51.82 & 162.34 & 0.521 & 56.78 & 77.34 & 0.502  & 54.23  & 119.97  \\ 
        1.0m & 0.536 & 54.13 & 114.23 & 0.459 & 51.93 & 162.46 & 0.523 & 56.89 & 79.27 & 0.506  & 54.32  & 118.65  \\ 
        1.5m  & \textbf{0.539} & \textbf{54.52} & \textbf{93.56} & \textbf{0.467} & \textbf{52.56} & \textbf{149.19} & \textbf{0.554} & 56.67 & 73.91 & \textbf{0.517} & \textbf{54.58} & \textbf{105.55} \\
        3.0m & 0.521 & 54.26 & 97.54 & 0.451 & 52.16 & 150.43 & 0.523 & \textbf{57.10} & \textbf{72.09} & 0.498  & 54.51  & 106.69  \\ 
        6.0m & 0.518 & 54.55 & 104.72 & 0.447 & 52.24 & 168.22 & 0.513 & 56.40 & 84.34 & 0.493  & 54.40  & 119.09 \\ 
\bottomrule
\end{tabular}}
\caption{Comparison of NTA-IoU, NTL-IoU, and FID scores for different stride settings in progressive data update strategy. Note that no progressive data update strategy is adopted when stride is set to 6m.}
\label{tab:incremental_integration_strategy}
\vspace{-1em}
\end{table*}
%###################################################

\section{Experiments}
In this section, we present our experimental setup, which encompasses the datasets, implementation details, and evaluation metrics. Subsequently, both quantitative and qualitative results are provided to demonstrate that the proposed \textit{ReconDreamer} can effectively render large maneuvers while also enhancing spatiotemporal coherence. Finally, we perform experiments to explore the different stride settings in progressive data update strategy and evaluate different \textit{DriveRestorer} backbone choices.

\subsection{Experiment Setup}

\noindent
\textbf{Dataset.} We conduct experiments in eight highly interactive scenes from the Waymo dataset \cite{waymo}. These scenes are characterized by numerous vehicles in various positions, following complex driving trajectories, with multiple lanes that increase the complexity of foreground and background reconstruction (see supplementary materials for specific scene IDs and frame numbers).
% We conduct experiments in the eight highly interactive scenes from the Waymo dataset. These scenes feature numerous vehicles in various positions, displaying complex driving trajectories. Additionally, the presence of multiple lanes adds to the complexity of foreground and background reconstruction. The supplementary materials provide specific scene IDs and frame numbers.
% We conduct experiments using the Waymo dataset\cite{waymo},  renowned for its extensive real-world driving data. However, most records capture relatively simple dynamic scenes that do not focus on intense vehicle interactions and involve smaller spaces, typically encompassing only one or two lanes. Eight scenes featuring high levels of interactive activity are selected to address this limitation, where numerous vehicles are in varied positions and exhibit complex driving trajectories. Additionally, these scenes include multiple lanes, making the foreground and background reconstruction more complex. Each of the chosen segments comprises around 40 frames, with their specific IDs provided in the supplementary materials.

\noindent
\textbf{Implementation Details.} To showcase the capability of \textit{ReconDreamer} in rendering large maneuvers, we conduct comprehensive comparisons against several state-of-the-art methods for dynamic driving scene reconstruction. These methods include Deformable-GS \cite{deformablegs}, $\text{S}^3$Gaussian \cite{s3gaussian}, PVG \cite{pvg}, Street Gaussians \cite{streetgaussian}, and DreamerDriver4D \cite{drivedreamer4d}. For the training phase, we divide the scenes into multiple segments, each containing 40 frames, and exclusively use data from the front-facing camera. Furthermore, we set up the training strategies and hyperparameters for each baseline method to match their original configurations, ensuring consistent training for 50,000 iterations. After the training, we assess the model's performance across three distinct novel trajectories: 3 meters lateral shift from the original trajectory, 6 meters lateral shift from the original trajectory, and lane change involving a lateral move to a parallel lane.

\noindent

\noindent
\textbf{Metrics.} 
Following DriveDreamer4D \cite{drivedreamer4d}, we use Novel Trajectory Agent IoU (NTA-IoU), Novel Trajectory Lane IoU (NTL-IoU), FID as evaluation metrics. Additionally, we conduct a user study to evaluate the quality of the rendered video. More details are in  supplementary materials.
\subsection{Main Results}
\noindent 
\textbf{Comparison with Scene Recosntruction Baselines.} In Tab.~\ref{tab:method_comparison}, we compare \textit{ReconDreamer} with different dynamic driving scene reconstruction methods \cite{pvg,s3gaussian,deformablegs,streetgaussian}. The experimental results demonstrate that the current state-of-the-art dynamic driving scene reconstruction method, Street Gaussians \cite{streetgaussian}, outperforms other traditional approaches (\ie, PVG \cite{pvg}, $\text{S}^3$Gaussian and Deformable-GS \cite{deformablegs}) across various novel trajectory renderings. Therefore, we investigate how much \textit{ReconDreamer} can improve performance based on \cite{streetgaussian}. Specifically, \textit{ReconDreamer} with Street Gaussians outperforms Street Gaussians across all metrics for different novel trajectory renderings. The average scores (NTA-IoU, NTL-IoU, FID) are relatively improved by 13.38\%, 3.35\%, and 26.40\%, respectively. Notably, \textit{ReconDreamer} with Street Gaussians effectively renders large maneuvers (e.g., Lane Shift @ 6m), relatively surpassing Street Gaussians by 24.87\%, 6.72\%, and 29.97\% on the NTA-IoU, NTL-IoU, and FID metrics, respectively.

\noindent \textbf{Comparison with DriveDreamer4D.} We compare the proposed \textit{ReconDreamer} with DriveDreamer4D\cite{drivedreamer4d}, both of which leverage world model priors to enhance driving scene reconstruction. Specifically, we implement both approaches on PVG\cite{pvg}, Deformable-GS\cite{deformablegs}, and $\text{S}^3$Gaussian\cite{s3gaussian}. The experimental results shown in Table \ref{tab:dd4d_comparison} indicate that these three original methods face challenges when rendering foreground vehicles and lane lines across new trajectories. While DriveDreamer4D improves over baseline algorithms in all metrics, its performance is limited in large maneuvers (6m lane shift), with NTA-IoU values of just 0.121, 0.031, and 0.078. 
In contrast, \textit{ReconDreamer}, which uses world model priors via an online restoration process instead of a training-free approach, shows superior performance compared to DriveDreamer4D. When compared to the versions of DriveDreamer4D implemented with PVG\cite{pvg}, $\text{S}^3$Gaussian\cite{s3gaussian}, and Deformable-GS\cite{deformablegs}, \textit{ReconDreamer} significantly boosts the average NTA-IoU scores by 44.00\%, 8.75\%, and 93.94\%, respectively. Furthermore, it achieves relative improvements in FID of 8.50\%, 28.93\%, and 35.39\%, respectively. Notably, in the large maneuver rendering (6m lane shift), ReconDreamer outperforms DriveDreamer4D, with relative improvements in NTA-IoU reaching 195.87\%, 470.97\%, and 276.92\%, respectively.

\noindent 
\textbf{User Study.} Additionally, a user study is conducted to evaluate the rendering quality of various methods on novel trajectories, with a particular focus on the presence of ghost artifacts in the rendered video. We select DriveDreamer4D with PVG \cite{drivedreamer4d} and Street Gaussians \cite{streetgaussian} as the comparison methods. As shown in Tab.~\ref{tab:user}, the results indicate that the \textit{ReconDreamer} approach significantly outperformed the above two methods in terms of user preference, which have an average of 97.22\% and 96.88\% win rates.

\begin{figure*}[!t]
\centering
\setlength{\abovecaptionskip}{0.2em}
\includegraphics[width=\textwidth]{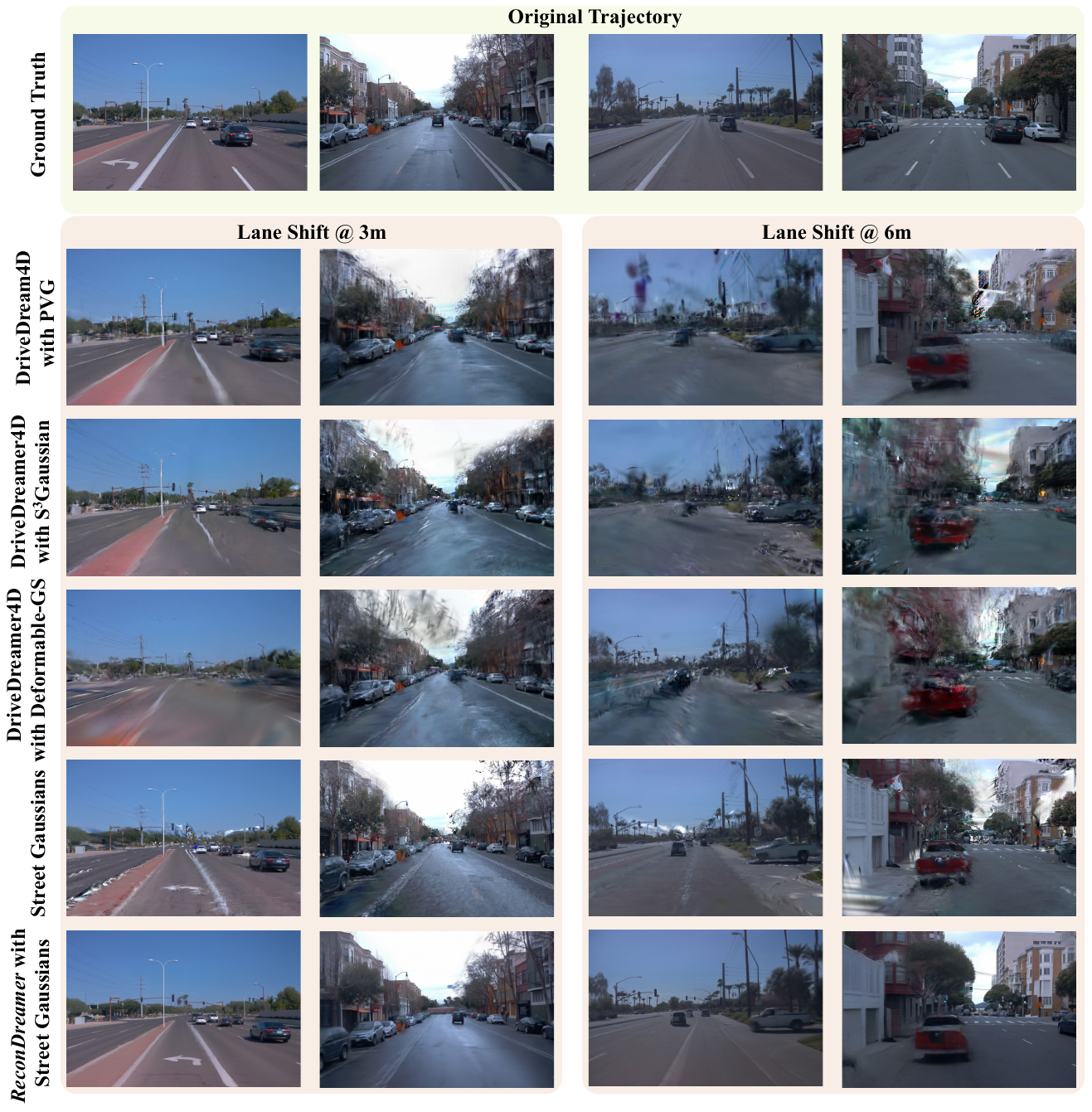}
\vspace{-0.2em}
\caption{Qualitative comparisons of novel trajectory renderings for lane shift @ 3m and lane shift @ 6m. The yellow box contains the ground truth video frames of the original trajectories, while the pink boxes display the rendered video frames after the lane shift.}
\label{fig:exp}
\vspace{-1.6em}
\end{figure*}

\noindent
\textbf{Qualitative Results.}
As illustrated in Fig.~\ref{fig:exp}, we present view renderings under new trajectories involving large viewpoint shifts, alongside their corresponding ground truth video from the original trajectories. 
The renderings from the baseline algorithms show significant ghosting and speckling in the background, with jagged lane markings and distorted or missing background elements such as trees. As the camera moves, numerous irregular white artifacts become visible, particularly in Street Gaussians \cite{streetgaussian}. Furthermore, foreground vehicles are severely distorted and affected by speckles. However, our method significantly improves the rendering quality. In the background, \textit{ReconDreamer} maintains excellent spatiotemporal consistency between the novel trajectory and the original trajectory. Additionally, the foreground vehicles remain clear and accurately reflect the viewpoint changes. 

% This improvement is particularly evident in the last frame (the rightmost column of Fig.~\ref{fig:exp}). In the baseline algorithms, this frame exhibits notable speckles and object distortion. In contrast, our method ensures a clean, consistent background and maintains clear, correctly positioned foreground vehicles.

\subsection{Ablation Study}

\noindent
\textbf{\textit{DriveRestorer} Backbone.}
As shown in Tab.~\ref{tab:pretrained}, we compare the performance of \textit{DriveRestorer} using different backbones and fine-tuning strategies, including Stable Diffusion \cite{sd}, Stable Video Diffusion \cite{svd}, DriveDreamer-2 \cite{drivedreamer2}, and DriveDreamer-2 with mask. All these methods utilize Street Gaussians \cite{streetgaussian} for reconstruction. The \textit{DriveRestorer} based on the Stable Diffusion \cite{sd} demonstrates strong performance, achieving relative improvements over the baseline by 2.41\%, 2.88\%, and 13.71\% in NTA-IoU, NTL-IoU, and FID metrics, respectively. However, image-based restoration models perform moderately on metrics such as NTA-IoU, primarily due to their lack of spatiotemporal consistency, which leads to positional deviation of vehicles after restoration. Meanwhile, the video-based method Stable Video Diffusion \cite{svd} provides better spatial continuity but faces challenges with high fine-tuning difficulty and lack of controllability, leading to color discrepancies and difficulties in restoring details such as lane lines. DriveDreamer-2 \cite{drivedreamer2}, built upon Stable Video Diffusion \cite{svd}, introduces additional control conditions, such as 3D boxes and HDMaps, significantly enhancing the NTA-IoU and NTL-IoU. Compared to Stable Diffusion \cite{sd}, DriveDreamer-2 improves the NTA-IoU by 9.42\% and the NTL-IoU by 0.13\%. Compared to Stable Video Diffusion \cite{svd}, the improvements are 4.93\% for NTA-IoU and 2.64\% for NTL-IoU. Additionally, the FID score improves by 7.13\% and 15.84\% compared to the respective baselines. Finally, incorporating masks during the fine-tuning process of DriveDreamer-2 \cite{drivedreamer2} further boosts the overall performance of the model. The supplement provides additional visual comparisons.

\noindent
\textbf{Progressive Data Update Strategy.} To analyze the influence of different stride settings ($\Delta y$ = 0.5, 1, 1.5, 3, and 6) in the PDUS, we conduct experiments on \textit{ReconDreamer} with Street Gaussians. As shown in Tab.~\ref{tab:incremental_integration_strategy}, the model achieves optimal performance when the stride is set to 1.5. Smaller stride values (0.5 and 1.0) lead to a 2.90\% and 2.13\% decrease in NTA-IoU, and a 0.64\% and 0.48\% reduction in NTL-IoU, respectively. Additionally, the FID score shows a significant drop of 13.66\% and 12.41\% respectively. Larger stride values cause excessive noise and ghosting during the rendering of novel trajectories, exceeding the restoration capacity of \textit{DriveRestorer}. Notably, a stride of 6 is equivalent to disabling the PDUS, resulting in the worst average NTA-IoU and NTL-IoU, which confirms the effectiveness of the proposed PDUS.
\section{Discussion and Conclusion}
\vspace{-0.4em}
% Closed-loop simulation plays a vital role in end-to-end autonomous driving, and accurate driving scene reconstruction is essential for achieving realistic simulations. While existing methods, such as NeRF and 3DGS, are effective at reconstructing scenes based on training data distributions, they struggle with rendering novel trajectories, particularly complex maneuvers like lane changes and multi-lane shifts. DriveDreamer4D has shown that integrating world model knowledge can alleviate some of these challenges, but it still faces limitations in rendering larger maneuvers.
% To address these shortcomings, we propose \textit{ReconDreamer}, which incrementally integrates world model knowledge to improve the driving scene reconstruction. By introducing \textit{DriveRestorer}, which mitigates ghosting artifacts through online restoration, and a progressive data update strategy, \textit{ReconDreamer} is the first method capable of rendering large maneuvers spanning up to 6 meters. Experimental results validate the superiority of \textit{ReconDreamer}, with substantial improvements over Street Gaussians in NTA-IoU, NTL-IoU, and FID, achieving relative improvements of 24.87\%, 6.72\%, and 29.97\%, respectively. Furthermore, \textit{ReconDreamer} outperforms DriveDreamer4D in large maneuver rendering, with a remarkable 195.87\% relative improvement in the NTA-IoU metric, as well as positive results from a comprehensive user study. These findings highlight the effectiveness of \textit{ReconDreamer} in enhancing dynamic driving scene reconstruction.
In closed-loop simulation, a crucial aspect is the ability to retrieve sensor data from any specified viewpoint, which can be achieved through accurate scene reconstruction. Methods like NeRF and 3DGS struggle with novel trajectory renderings. DriveDreamer4D alleviates the challenge by using a pre-trained world model but still struggles with larger maneuvers. To address these limitations, we propose \textit{ReconDreamer}, a system designed to enhance driving scene reconstruction via online restoration. \textit{ReconDreamer} incorporates \textit{DriveRestorer} to effectively reduce artifacts and leverages a progressive data update strategy, ensuring high-quality rendering even for large-scale maneuvers spanning up to 6 meters. Experiments show \textit{ReconDreamer} surpasses Street Gaussians in NTA-IoU, NTL-IoU, and FID with improvements of 24.87\%, 6.72\%, and 29.97\%, respectively. It also outperforms DriveDreamer4D with PVG in large maneuver rendering, with a comprehensive user study and 195.87\% improvement in NTA-IoU. 
%These results demonstrate the effectiveness of \textit{ReconDreamer} in advancing dynamic driving scene reconstruction.

{
    \small
    \bibliographystyle{ieeenat_fullname}
    \bibliography{PaperForReview}
}


\clearpage

\twocolumn[{%
% \vspace{-1em}
\maketitle
% \vspace{-2em}
\begin{center}
% \vspace{-2em}
\centering
\includegraphics[width=\textwidth]{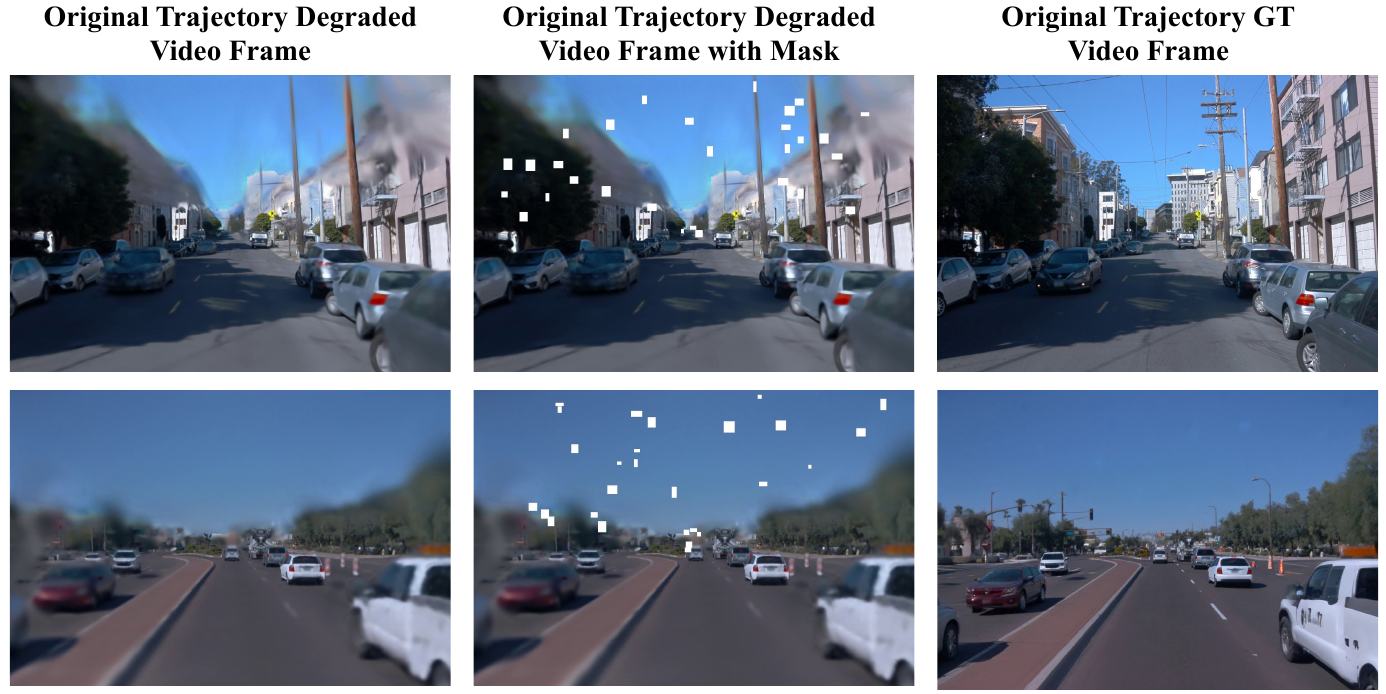}
\captionof{figure}{
Examples of degraded video frames, the corresponding masked video frames, and GT video frames under the original trajectories.}
\label{fig:mask}
\end{center}}]

\begin{figure*}[!ht] 
\centering
\includegraphics[width=\textwidth]{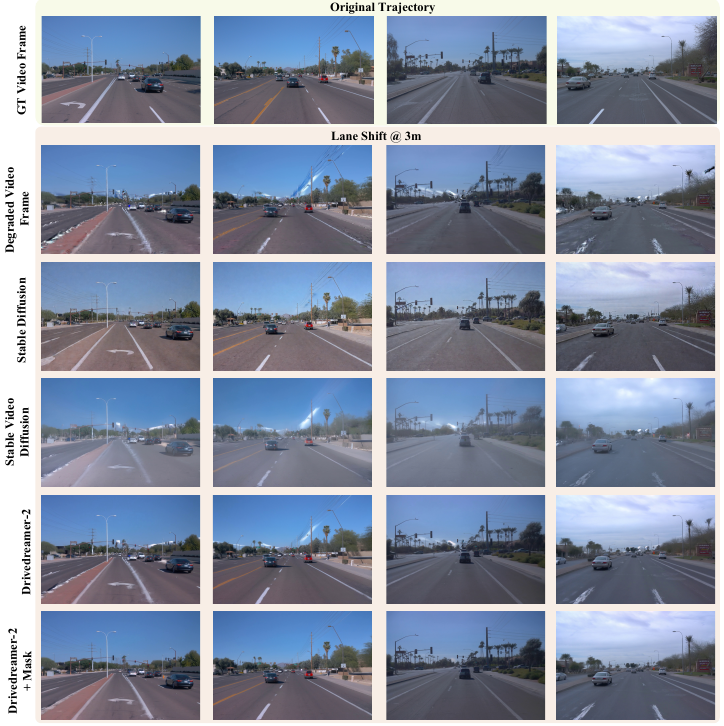}
\caption{Qualitative comparison of the restoration effects achieved by \textit{DriveRestorer}  with different backbones. The yellow box contains the ground truth video frames of the original trajectories, while the pink boxes display the rendered video frames after the lane shift and the corresponding restored video frames by \textit{DriveRestorer}  with different backbones.
}
\label{fig:backbone}
%\vspace{-1.5em}
\end{figure*}

In the supplementary material, we provide detailed information on the training for \textit{DriveRestorer} , the selected scenes, the metrics, and the user study. Additionally, we present more qualitative results to compare the restoration effects achieved by \textit{DriveRestorer}  with different backbones and to evaluate the impact of varying stride settings in the PDUS.

\section{Implementation Details}

\noindent \textbf{Training for \textit{DriveRestorer}.} As shown in Fig.~\ref{fig:mask},the frames rendered by the reconstruction model often exhibit significant degradation at the boundary between the sky and the background and the areas far from the camera in the image center. To address these issues, we introduce a masking strategy, applying random masks to these degraded regions to guide the model in repairing them. 

\noindent \textbf{Metrics.} As mentioned in the main text, we utilize Novel Trajectory Agent Intersection over Union (NTA-IoU) and Novel Trajectory Lane Intersection over Union (NTL-IoU) to assess the quality of the rendered video, both metrics proposed in DriveDreamer4D \cite{drivedreamer4d}. These metrics are specifically designed to evaluate the spatiotemporal coherence of foreground agents and background lanes, respectively.

The NTA-IoU processes images rendered under new trajectories using the YOLO11 \cite{yolo11_ultralytics} detector to extract 2D bounding boxes of vehicles. Meanwhile, by applying geometric transformations to the 3D bounding boxes from the original trajectories, they can be accurately projected onto the new trajectory perspective, thus obtaining the ground truth 2D bounding boxes in the new trajectory view. Each projected 2D bounding box will find the nearest 2D bounding box generated by the detector and compute their Intersection over Union (IoU). 

Similarly, the NTL-IoU employs the TwinLiteNet \cite{che2023twinlitenet} model to detect lane in the images rendered under the new trajectories, and the lane from the original trajectories will also be projected onto the new trajectory through corresponding geometric transformations.  Finally, the mean Intersection over Union (IoU) between the projected and detected lane lines is calculated.

\noindent \textbf{Scene Selection.} 
We select eight scenes from the validation set of the Waymo dataset \cite{waymo}. These scenes feature high levels of interactive activity, with numerous vehicles in varied positions and exhibiting complex driving trajectories. Additionally, these scenes include multiple lanes, which increases the complexity of foreground and background reconstruction. As shown in Table. \ref{tab:scene}, we provide a detailed list of the segment IDs.

\noindent \textbf{User Study.} In the user study, we compare our results with two baseline models: DriveDreamer4D with PVG \cite{drivedreamer4d} and Street Gaussians \cite{streetgaussian}. This comparison is conducted across the eight scenarios we selected, with an emphasis on the overall quality of the videos, including the consistency and clarity of the background, as well as the positional accuracy of foreground objects. In each comparison, our method and the baseline methods are randomly assigned to the top or bottom of the video, and participants are asked to choose the option they find most satisfactory.

\begin{table*}
    \centering
    \begin{tabular}{lcc}
        \toprule
        \makecell[c]{Scene} & Start Frame & End Frame \\ \midrule
    segment-10359308928573410754\_720\_000\_740\_000\_with\_camera\_labels.tfrecord & 120 & 159 \\
    segment-11450298750351730790\_1431\_750\_1451\_750\_with\_camera\_labels.tfrecord & 0 & 39 \\
    segment-12496433400137459534\_120\_000\_140\_000\_with\_camera\_labels.tfrecord & 110 & 149 \\
    segment-15021599536622641101\_556\_150\_576\_150\_with\_camera\_labels.tfrecord & 0 & 39 \\
    segment-16767575238225610271\_5185\_000\_5205\_000\_with\_camera\_labels.tfrecord & 0 & 39 \\
    segment-17860546506509760757\_6040\_000\_6060\_000\_with\_camera\_labels.tfrecord & 90 & 129 \\
    segment-3015436519694987712\_1300\_000\_1320\_000\_with\_camera\_labels.tfrecord & 40 & 79 \\
    segment-6637600600814023975\_2235\_000\_2255\_000\_with\_camera\_labels.tfrecord & 70 & 109 \\
        \bottomrule
        \end{tabular}
        \caption{Eight scenes from the Waymo dataset \cite{waymo} featuring high interactive activity, numerous vehicles, and complex driving trajectories.}
        \label{tab:scene}
\end{table*}

\begin{figure*}[!ht] 
\centering
\includegraphics[width=\textwidth]{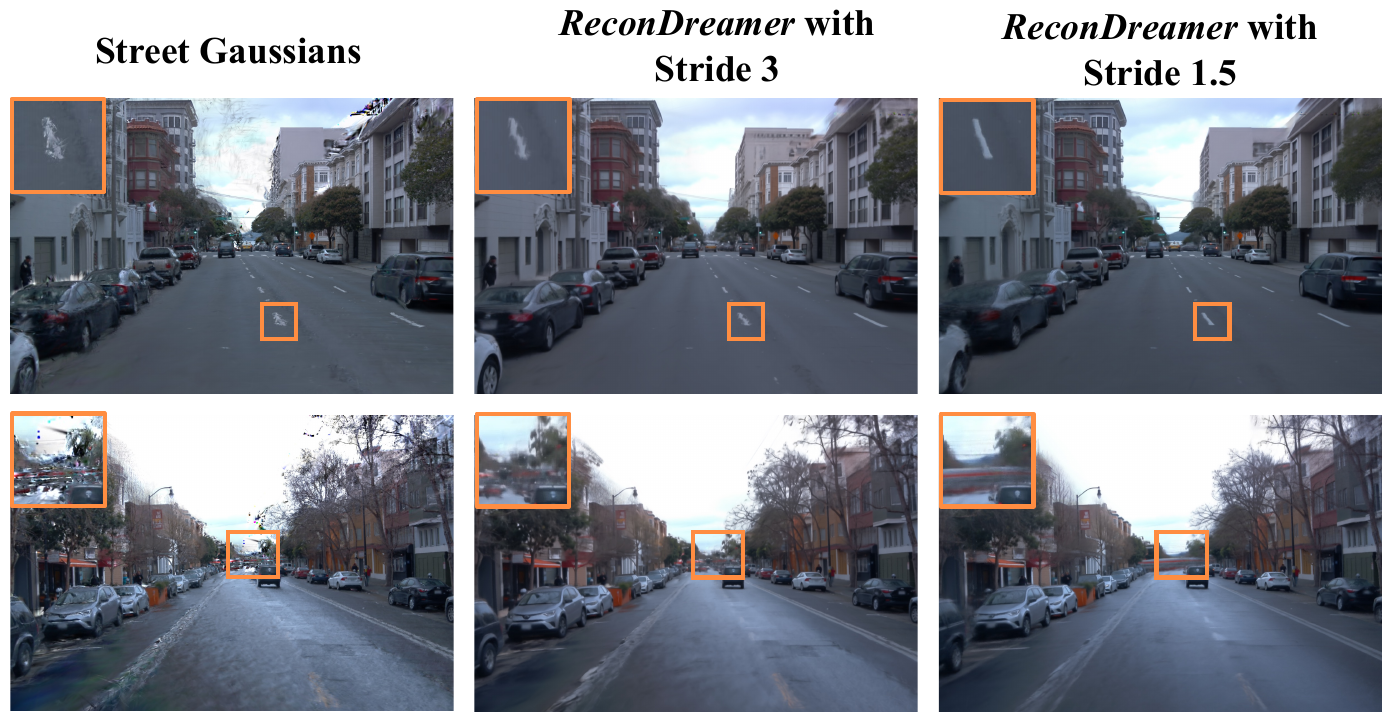}
\caption{
 Qualitative comparison of the different stride settings in the PDUS.}
\label{fig:stride}
%\vspace{-1.5em}
\end{figure*}

\section{Baseline}
\noindent \textbf{PVG} \cite{pvg}  introduces a novel unified representation model designed to capture dynamic scenes through the use of time-varying Gaussian distributions. These Gaussians are characterized by adjustable properties such as vibration direction, duration, and peak intensity. The approach distinguishes between static and dynamic elements by sorting the Gaussians according to their durations. 

\noindent \textbf{Deformable-GS} \cite{deformablegs} establishes a canonical space where scenes are represented using Gaussian distributions. For capturing dynamics, it employs a deformation network to forecast the offsets of Gaussian attributes, which subsequently adjust the Gaussians to align with the scene's dynamic changes

\noindent \textbf{$\text{S}^3$Gaussian} \cite{s3gaussian} is a method designed for efficient 3D scene reconstruction that operates without the need for expensive annotations. It achieves this by using 4D consistency to divide scenes into dynamic and static components, representing each with 3D Gaussians for detailed precision and employing a spatial-temporal field network to model the 4D dynamics compactly.

\noindent \textbf{Street Gaussians} \cite{streetgaussian} is a dynamic scene modeling method based on Gaussian Splatting for driving scenes. It separately models the static background and foreground vehicles. By utilizing boxes predicted by a pre-trained model, Street Gaussians warps the Gaussians of foreground vehicles and refines them during training.

\noindent \textbf{DriveDreamer4D} \cite{drivedreamer4d} is a method that enhances dynamic driving scene reconstruction by integrating with state-of-the-art techniques such as PVG \cite{pvg}, Deformable-GS \cite{deformablegs}, and $\text{S}^3$Gaussian \cite{s3gaussian}. It leverages world model priors to synthesize novel trajectory videos, where structured conditions are explicitly utilized to control the spatial-temporal consistency of traffic elements.

\section{Experiment Results}
\noindent \textbf{Qualitative Results of \textit{DriveRestorer} Backbone.}
As shown in Fig. \ref{fig:backbone}, we compare restoration effects achieved by \textit{DriveRestorer}  with different backbones. 
The images rendered under the new trajectories exhibit several defects, including distorted and blurred distant trees, flying points in the sky, and partially obscured foreground vehicles. The \textit{DriveRestorer}  based on Stable Diffusion  \cite{sd} demonstrates promising performance, repairing the background and effectively correcting the distortion of foreground vehicles. However, image restoration methods lack spatial continuity, causing the repaired foreground vehicles to appear in incorrect positions or even exhibit color changes. For instance, in the second column, some distant vehicles that are originally red turned into grey. The video-based method, Stable Video Diffusion \cite{svd}, offers improved spatial continuity but encounters challenges due to the great difficulty of fine-tuning. Although it restores many distorted vehicles, the video frames show significant color differences from the original and sky defects remain unrepaired. Then, DriveDreamer-2 \cite{drivedreamer2} introduces control conditions, such as 3D boxes and HDMaps, which resolve the issue of color discrepancies and improve the restoration of background elements like lane lines. Finally, incorporating masks during the fine-tuning process of DriveDreamer-2 \cite{drivedreamer2} further enhances the repair of sky defects, making the restored video frame more realistic.

\noindent \textbf{Qualitative Results of Progressive Data Update Strategy.} In Figure~\ref{fig:stride}, we compare different stride settings in the PDUS, including $\Delta y = 1.5$ and $3$. Although \textit{ReconDreamer} is effective in enhancing image quality and reducing artifacts for both stride values, an excessively large stride can lead to poorer reconstruction of lane markings and distant scenes.

We provide a video that includes more comparisons with the baseline. For further details, please refer to the file videos/comparison.mp4.

\end{document}